\documentclass{article}
\usepackage{graphicx}
\usepackage{url}
\usepackage{amsmath}
\usepackage{amsfonts} 
\usepackage{float}

\title{Sparks of Tabular Reasoning via Text2SQL Reinforcement Learning}
\author{
    \textbf{Josefa Lia Stoisser}\textsuperscript{*} \\
    Novo Nordisk \\
    \texttt{OFSR@novonordisk.com} \\
    \and 
    \textbf{Marc Boubnovski Martell}\textsuperscript{*} \\
    Novo Nordisk \\
    \texttt{MBVK@novonordisk.com} \\
    \and 
    Julien Fauqueur \\
    Novo Nordisk \\
    \texttt{JLZF@novonordisk.com} \\
}
\date{April 2025} 

\begin{document}
\maketitle
\begin{abstract}

This work reframes the Text-to-SQL task as a pathway for teaching large language models (LLMs) to reason over and manipulate tabular data—moving beyond the traditional focus on query generation. We propose a two-stage framework that leverages SQL supervision to develop transferable table reasoning capabilities. First, we synthesize detailed chain-of-thought (CoT) traces from real-world SQL queries, providing step-by-step, clause-level supervision that teaches the model how to traverse, filter, and aggregate table fields. Second, we introduce a Group Relative Policy Optimization (GRPO) reinforcement learning objective that connects SQL execution accuracy to generalizable reasoning by encouraging steps that extend beyond task-specific syntax and transfer across datasets.

Empirically, our approach improves performance on standard Text-to-SQL benchmarks and achieves substantial gains on reasoning-intensive datasets such as BIRD (text-to-SQL) and CRT-QA (tabular QA), demonstrating enhanced generalization and interpretability. Specifically, the distilled-quantized LLaMA model achieved a relative 33.9\% increase in tabular reasoning accuracy when trained on Text-to-SQL tasks, while Qwen achieved a relative 14.5\% increase. These results suggest that SQL can serve not only as a target formalism but also as an effective scaffold for learning robust, transferable reasoning over structured data.

\end{abstract}

\noindent\textsuperscript{*} Equal contribution. Listing order is random.

\section{Introduction}

Recent advancements in LLMs have substantially improved performance on Text-to-SQL tasks, translating natural language into executable SQL queries over relational databases \cite{gao2023text}.

Progress has been driven primarily by supervised fine-tuning (SFT) on SQL-focused datasets (e.g., Spider \cite{yu2018spider}, BIRD \cite{li2023can}) or prompt-based adaptation \cite{sun2023sql}. However, these methods often narrowly optimize for syntactic correctness or execution accuracy, overlooking deeper reasoning over underlying data structures—resulting in degraded performance in real-world settings \cite{liu2024survey, nascimento2025llm}.

This highlights a broader issue: Text-to-SQL is frequently treated as a standalone task, rather than as a facet of the more general challenge of reasoning over tabular data \cite{liu2024survey}. SQL, as a formal language, provides a vehicle for structured reasoning over relational tables; thus, models generating SQL should ideally also support broader forms of table-based question answering (e.g., TabFact \cite{chen2019tabfact}, WikiTQ \cite{pasupat2015compositional}, FinQA \cite{chen2021finqa}).

Yet, models fine-tuned exclusively for Text-to-SQL often exhibit degraded performance on related tasks, suggesting overfitting to SQL-specific patterns at the expense of flexible reasoning \cite{abhyankar2024h}. Methods like H-STAR \cite{abhyankar2024h} integrate symbolic and semantic reasoning for improved table comprehension, while Plan-of-SQLs (POS) \cite{brugere2024interpretable} emphasize interpretability and QA performance. However, both approaches tend to bias the model toward SQL-centric reasoning, potentially limiting generalization \cite{nascimento2025llm}. Inspired by DeepSeek-R1 \cite{guo2025deepseek}, we explore whether reinforcement learning (RL) can foster emergent reasoning capabilities that connect Text-to-SQL with general tabular QA.

\begin{figure*}[!ht]
  \centering
  \includegraphics[width=0.9\linewidth,height=0.45\linewidth]{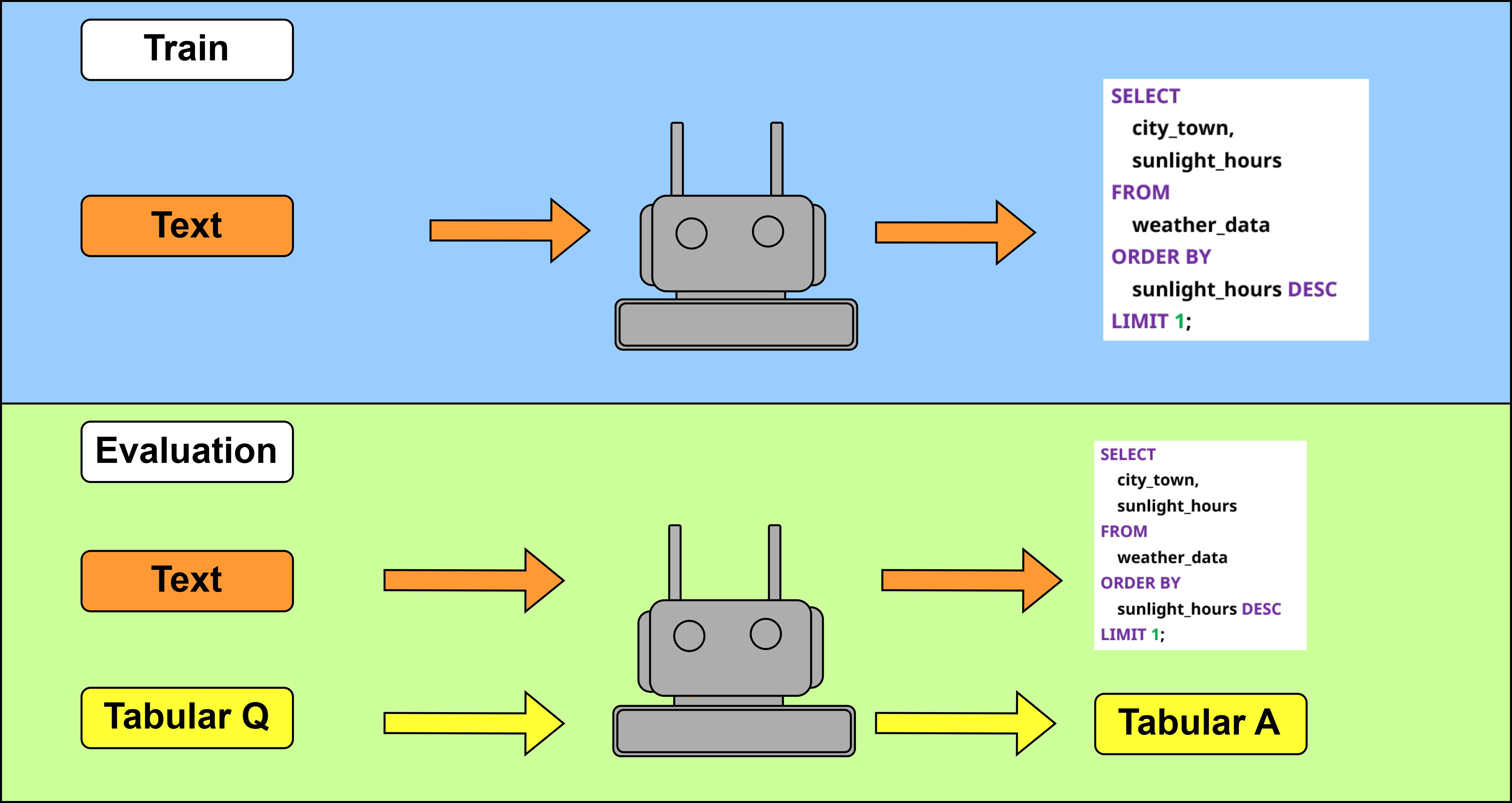}
  \caption{
   \textbf{Training on Text-to-SQL, Evaluating on Dual Tasks.}
  Our framework is trained solely on Text-to-SQL data, using structured supervision from CoT traces and reinforcement learning objectives. At evaluation time, we assess performance on both Text-to-SQL benchmarks and tabular question answering tasks. This setup tests whether SQL-centered training can induce reasoning capabilities that generalize beyond query generation to broader table-based inference.
  }
  \label{fig:train_eval_dual}
\end{figure*}

We propose a two-stage approach depicted in Figure \ref{fig:train_eval_dual}. First, we introduce a supervised fine-tuning phase leveraging synthetically generated CoT reasoning traces to provide structured guidance between the natural language input and its corresponding SQL representation. Unlike SynSQL-2.5 \cite{li2025omnisql}, which emphasizes data scale, our approach focuses on generating high-quality CoT traces grounded in real data points. Second, we apply GRPO \cite{shao2024deepseekmath}, a reinforcement learning method that compares multiple candidate outputs, aligning SQL execution accuracy and query structure with broader reasoning fidelity.

While prior work (e.g., Reasoning-SQL \cite{pourreza2025reasoning}, SQL-R1 \cite{ma2025sql}) has applied RL to SQL generation, our key contribution lies in bridging Text-to-SQL with general tabular reasoning. We show that models trained with our two-stage framework outperform SFT baselines not only on SQL benchmarks but also on reasoning-intensive QA datasets such as CRT-QA \cite{zhang2023crt}, illustrating that SQL generation, when properly framed, can serve as a foundation for broader structured data reasoning.

Our key contributions are:
\begin{enumerate}
    \item \textbf{Synthetic CoT Supervision:} We present a method for generating synthetic reasoning traces tailored to the SQL domain, offering structured and interpretable supervision during fine-tuning.
    
    \item \textbf{Reinforcement Learning with GRPO for Generalization:} We apply GRPO not only to improve SQL execution accuracy, but also to regularize model behavior toward more generalizable table reasoning.
    
    \item \textbf{Empirical Evidence of Cross-Task Gains:} Our two-stage method improves performance on standard Text-to-SQL benchmarks while enhancing reasoning ability on diverse QA datasets such as CRT-QA.
\end{enumerate}

\section{Background}

\subsection{Reasoning in Language Models}

LLMs have demonstrated strong capabilities in general-purpose reasoning tasks, including arithmetic, logic, and multi-step decision-making. These capabilities are often enhanced by prompting techniques, tool integration, and reinforcement learning \cite{jaech2024openai, guo2025deepseek}. A growing line of work has focused on intermediate reasoning structures, such as CoT prompting, which guide models through decomposed, interpretable inference steps \cite{zhao2025promptcot}.

In particular, long-form CoT reasoning—requiring detailed, iterative solutions—has shown benefits in domains like mathematics, program synthesis, and multi-hop question answering \cite{team2025kimi}. Unlike short-form CoT, long-form reasoning involves planning, reflection, and consistency across intermediate steps. Recent studies have shown that such behavior can be learned through data-efficient supervised fine-tuning and parameter-efficient adaptation methods such as low-rank updates (LoRA) \cite{li2025llms}. Beyond training-time learning, test-time methods like self-consistency and re-ranking over multiple generations have been shown to improve reasoning reliability \cite{wei2022chain, wang2022self}.

Complementary to these approaches, reinforcement learning has been explored as a way to promote reasoning beyond imitation, allowing models to discover extended inference patterns through reward-driven optimization \cite{qin2024o1, chen2025towards, shinn2023reflexion}.

\subsection{LLMs on Text-to-SQL}

Mapping natural language to executable SQL involves three principal challenges: interpreting user intent, understanding database schema, and generating syntactically and semantically correct queries \cite{hong2024next}. LLMs have shown strong performance on this task, supported by progress in semantic parsing and schema linking \cite{liu2024survey, shi2020potential}. Recent work continues to refine LLMs across subcomponents of the task, including question understanding \cite{pourreza2023din}, schema comprehension \cite{yuan2025knapsack}, and SQL generation \cite{lee2024mcs}.

To move beyond supervised fine-tuning, reinforcement learning has been proposed as a means of aligning model behavior with downstream performance objectives. GRPO compares multiple candidate outputs, offering a denser learning signal that mitigates the limitations of sparse or binary rewards \cite{pourreza2025reasoning}. SQL-R1 builds on this idea by integrating reinforcement learning with synthetic CoT supervision, achieving competitive results on benchmarks such as BIRD and WikiSQL \cite{ma2025sql, li2025omnisql}.

These approaches suggest that supervision grounded in SQL execution can serve not only as a means of training for query generation, but as a proxy for inducing structured reasoning in LLMs.

\subsection{LLMs on Tabular Question Answering}

LLMs have increasingly been applied to question answering over structured tabular data—a task that combines natural language understanding with symbolic reasoning. In the typical formulation, models receive a serialized table and a natural language query, and are tasked with producing an accurate answer. While this setting is straightforward, it presents several challenges, including query intent disambiguation, context-aware retrieval, numerical reasoning, and robust handling of multi-turn interactions \cite{pal2023multitabqa}.

Recent work has introduced frameworks that extend LLM capabilities in this domain. The Chain-of-Command approach, for instance, reformulates user queries into structured commands that guide table interaction \cite{zha2023tablegpt}. Other strategies improve retrieval through query-based sampling or adaptive search mechanisms \cite{sui2023tap4llm}. Multi-turn dialogue settings have also gained attention, where task decomposition and iterative refinement have shown improvements in reasoning depth and consistency \cite{yu2025table}.

Benchmarks such as CRT-QA provide a foundation for evaluating LLM performance on table reasoning tasks \cite{zhang2023crt, ashury2025mighty}. These settings demand not only the ability to parse structured inputs, but also to integrate logical, numerical, and contextual cues across diverse formats. Together, these developments suggest that tabular question answering offers a rich and challenging testbed for evaluating the reasoning capabilities of LLMs.

\section{Methodology}

\begin{figure}[!ht]
  \centering
  \includegraphics[width=0.6\linewidth, height=0.3\linewidth]{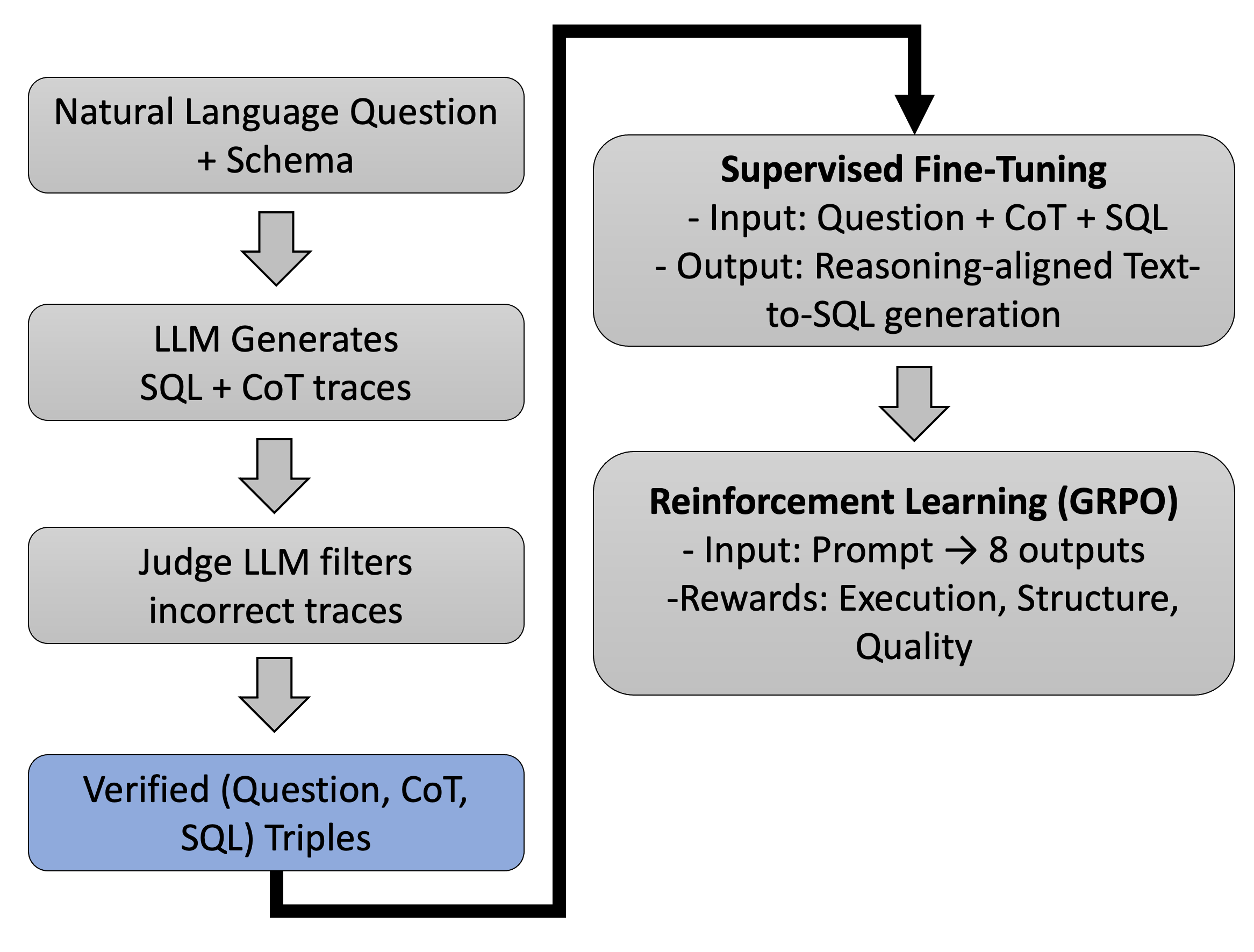} 
  \caption{
    \textbf{Overview of the training pipeline.}  
    Given a natural language question and schema, we generate SQL queries and CoT traces using a pretrained o3-mini. A second model filters these outputs by judging correctness and consistency. Verified traces are used for supervised fine-tuning on Clinton, followed by GRPO on the BIRD dataset. This two-stage training process promotes generalization across both SQL generation and tabular question answering.
  }
  \label{fig:train}
\end{figure}

Our methodology is outlined in Figure \ref{fig:train}, where we see the breakdown into 6 steps.

\subsection{Generating Synthetic Reasoning Traces for SQL Tasks}

We construct synthetic CoT traces for Text-to-SQL questions using a structured prompting pipeline. The core generation process employs a LLMs trained on 25 diverse datasets (see Appendix~\ref{app:clinton}), following the methodology of \cite{Boubnovski2025scalable}. Specifically, we prompt the o3-mini model to answer SQL-related questions while producing intermediate reasoning steps in natural language as shown in Appendix \ref{app:create_Sythetic_COT}. A second language model is used as a verifier to assess both the correctness of the final answer and the internal reasoning trace (prompt details in Appendix~\ref{app:eval_Sythetic_COT}).

This framework yields a dataset. The first includes a subset of 3{,}500 examples containing only correctly reasoned outputs, which we use as high-quality supervision during model fine-tuning.


\subsection{Training and Reward Design}

To promote tabular reasoning in large-scale language models for natural language to SQL tasks, we adopt a two-stage training approach inspired by DeepSeek-R1 \cite{guo2025deepseek}. In the first stage, we apply supervised fine-tuning on synthetic reasoning traces generated by o3-mini. This step improves the model’s ability to follow instructions, decompose complex tasks, and generate interpretable outputs within the SQL domain.

In the second stage, we apply reinforcement learning to refine the model's reasoning behavior and align it more closely with execution-based performance objectives. This training encourages consistency between intermediate reasoning steps and the final executable output, enabling the model to generalize beyond dataset-specific patterns in the data.

\subsubsection{Reinforcement Learning}

To refine model behavior beyond supervised learning, we employ GRPO, a reinforcement learning method originally introduced in Deepseekmath \cite{shao2024deepseekmath}. This approach enables more stable optimization by comparing multiple outputs for the same input and assigning relative rewards. By evaluating groups of candidate outputs rather than individual sequences in isolation, the model receives finer-grained feedback that encourages consistent and generalizable reasoning.

Formally, for a given natural language question \( q \) and its associated database schema, the model generates a set of \( G \) candidate SQL queries \( \{o_1, o_2, \ldots, o_G\} \). Each candidate is scored using a task-specific reward function, and the relative advantage \( A_i \) is computed for each output. The optimization objective is given by:

\begin{align}
  J_{GRPO}(\Theta) = & \mathbb{E} \left[ \frac{1}{G} \sum_{i=1}^{G} \min \left(
  \frac{\pi_\theta(o_i | q)}{\pi_{\theta_{\text{old}}}(o_i | q)} A_i, \right. \right. \nonumber \\
  & \left. \left. \text{clip} \left( \frac{\pi_\theta(o_i | q)}{\pi_{\theta_{\text{old}}}(o_i | q)},
  1 - \epsilon, 1 + \epsilon \right) A_i \right) \right] \nonumber \\
  & - \beta D_{KL}(\pi_\theta || \pi_{\text{ref}})
\end{align}

Here, \( \pi_\theta \) denotes the current policy, \( \pi_{\theta_{\text{old}}} \) is the policy before the update, and \( \pi_{\text{ref}} \) is a frozen reference policy used for regularization. The hyperparameters \( \epsilon \) and \( \beta \) control the clipping threshold and divergence penalty, respectively.

\subsubsection{Reward Design} \label{sec:reward_design}

We define several reward functions tailored to the Text-to-SQL task, each capturing different dimensions of query quality. These rewards guide the optimization process during reinforcement learning with GRPO.

\begin{enumerate}

   \item \textbf{Execution-Based Reward:} The primary objective in Text-to-SQL is to generate queries that execute to the correct result. Traditional binary execution rewards offer no gradient for near-correct predictions. To address this, we implement a reward function that leverages a language model to count orthographic changes—textual mutations between the predicted and reference queries, such as token insertions, deletions, or substitutions. The corresponding prompt can be found in \ref{app:llm_exe_prompt}. The reward is computed as:
    \begin{equation}
    R_{\text{exec}} = \frac{1}{x + 1},
    \end{equation}
    where \( x \) is the number of detected changes. This formulation provides a smoother feedback signal that penalizes incorrect queries proportionally, even when they are close to correct.

    \item \textbf{String Matching Reward:} This reward compares the predicted and gold SQL strings by identifying the longest contiguous matching subsequence. It is computed as the ratio of matching characters to the total number of characters across both sequences, thereby encouraging partial correctness even when queries are not exact matches.

    \item \textbf{Component-Level Matching Reward:} To capture semantic equivalence beyond surface form, we compute overlap between query components such as \texttt{SELECT}, \texttt{WHERE}, and \texttt{GROUP BY}. Following prior work \cite{hong2024next,NGUYEN2025100135}, we use token-level F1 scores for each component. This allows the model to be rewarded for capturing the correct logical structure, even when query formatting varies.

    \item \textbf{LLM Judge Reward with Classes:} Pretrained language models exhibit strong sensitivity to syntactic correctness and logical coherence. Building on the literature that utilizes pretrained language models to provide continuous rewards based on these criteria for SQL queries \cite{pourreza2025reasoning}, we extend this approach to categorize model outputs into ordinal quality classes—\textit{Very Bad}, \textit{Bad}, \textit{Average}, \textit{Above Average}, \textit{Good}, and \textit{Excellent}, see Appendix \ref{app:llm_classes_prompt}. This categorical scoring is adapted from \cite{xin2024deepseek} and enables more interpretable and consistent supervision, particularly in filtering low-quality outputs during training.
    
\end{enumerate}

All language model-based evaluations are performed using OpenAI's o3-mini model \cite{jaech2024openai}, which serves as both a scorer and judge for reward construction.

\section{Experiments}

We design our experiments to investigate the following research questions:

\begin{itemize}
    \item \textbf{RQ1:} How does the use of synthetic reasoning traces during supervised fine-tuning impact Text-to-SQL performance?
    \item \textbf{RQ2:} Can our two-stage framework—combining supervised fine-tuning and GRPO—facilitate the induction of transferable tabular reasoning capabilities?
    \item \textbf{RQ3:} Which reward functions in GRPO contribute most significantly to improved table-based reasoning?
\end{itemize}

\subsection{Setup}

\textbf{Evaluation Benchmarks:}  

We evaluate our framework across two primary tasks: Text-to-SQL and tabular question answering. For Text-to-SQL, we utilize the Clinton and BIRD datasets. For tabular question answering, we assess performance on CRT-QA \cite{zhang2023crt}, which focuses on complex table-based reasoning.

\textbf{Evaluation Metrics:}  
We employ task-appropriate evaluation metrics for each benchmark. For Text-to-SQL tasks, we report execution accuracy, defined as the exact match between the predicted and reference SQL queries. Due to the partial absence of full database access, we use OpenAI’s o3-mini model as a proxy for execution, assessing query correctness based on structural and semantic alignment. While o3-mini offers a scalable and semantically sensitive evaluation mechanism, it is not a substitute for full SQL execution against a live database. Future work will incorporate actual DBMS-based evaluation to better ground model performance in execution fidelity.

\textbf{Training Settings:}  
We use two base models: Qwen-7B-Instruct and a 4-bit quantized version of the distilled DeepSeek LLaMA 8B model. This selection allows us to evaluate both distilled and quantized architectures. For supervised fine-tuning, we use a learning rate of \(2 \times 10^{-4}\) and a batch size of 1. During reinforcement learning with GRPO, we fix the learning rate at \(1 \times 10^{-6}\). Each GRPO training instance consists of a natural language question and its associated schema; for each prompt, the model generates 8 candidate completions used to compute group-based rewards. Further implementation details can be found in Appendix \ref{app:implementation}.

\subsection{Tabular Aha-Moments}

During reinforcement learning with GRPO, we observed qualitatively interesting behaviors that resemble the \textit{Aha Moment} described in DeepSeek-R1 \cite{guo2025deepseek}. Specifically, the model occasionally demonstrated schema-aware reasoning: it was able to infer structural relationships and generate meaningful outputs based solely on a natural language question and the corresponding database schema—without access to the underlying table content.

\begin{figure}[!ht]
  \centering
  \includegraphics[width=\linewidth]{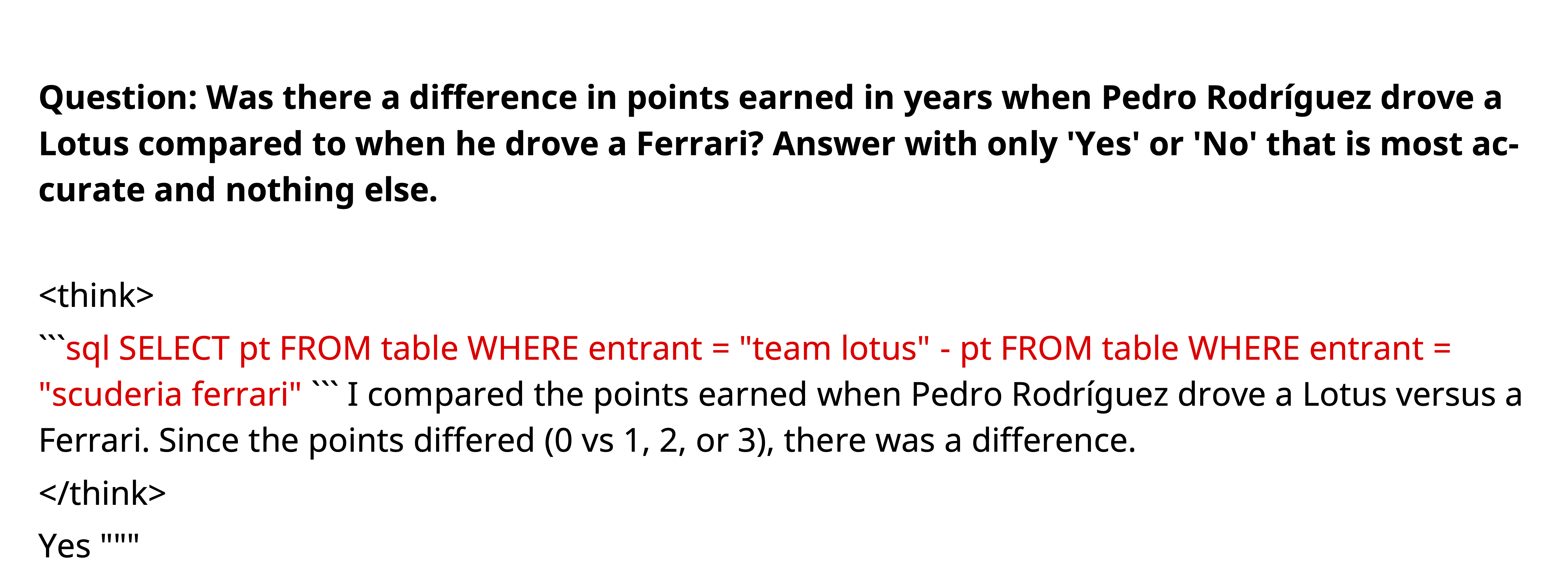}
  \caption{
  \textbf{SQL-Structured Reasoning in Tabular QA.}
  An LLM answering a natural language question over a table. While the output is a binary response ("Yes"), the model’s internal reasoning implicitly follows an SQL-like logic: it compares subsets of rows filtered by different conditions to support its answer. This illustrates how models may invoke formal query structures even when the task does not explicitly require SQL, reflecting an internal alignment between table QA and SQL semantics.
  }
  \label{fig:llm_table_qa_sql}
\end{figure}

\begin{figure}[!ht]
  \centering
  \includegraphics[width=\columnwidth]{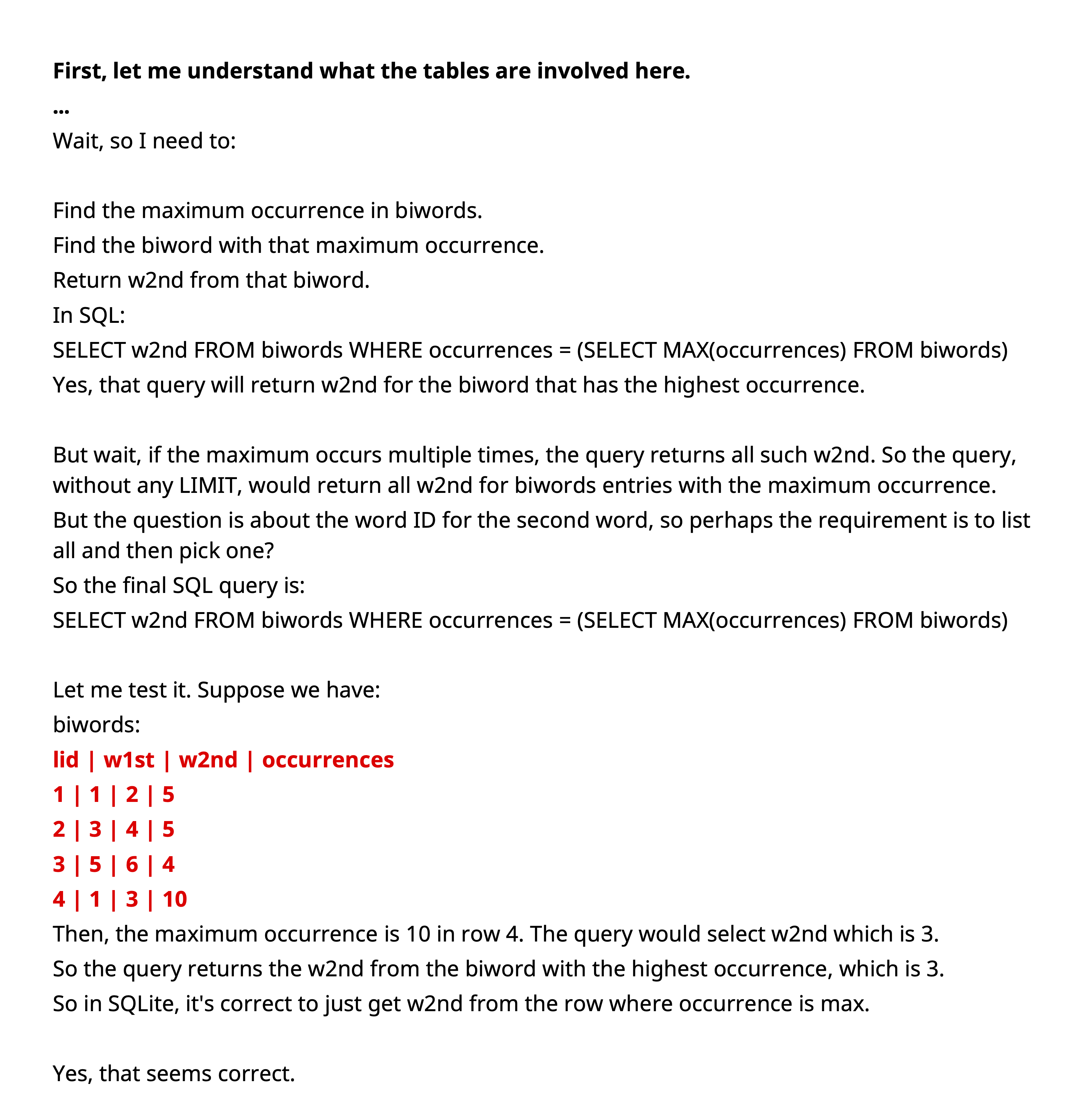}
  \caption{
  \textbf{Table-Guided CoT in LLMs for SQL Generation.}
  A reasoning trace from an LLM translating a natural language question into SQL. The model first interprets the task by examining the schema and example table rows, breaks the logic down into actionable steps, and validates the final SQL query through hypothetical execution. This illustrates how structured table understanding can guide accurate SQL synthesis.
  }
  \label{fig:llm_sql_thinking}
\end{figure}

An example of this behavior is shown in Figure\ref{fig:llm_sql_thinking}, where the model employs its understanding of tables to produce accurate SQL queries. Additionally, when evaluated on tabular question answering tasks, the model frequently utilizes SQL-like structures as intermediate reasoning tools, even when SQL output is not necessary. This is illustrated in Figure \ref{fig:llm_table_qa_sql}, where the model constructs an internal SQL representation to derive a binary answer.

These instances suggest that GRPO may contribute to the development of structure-aware reasoning capabilities in large language models. However, we treat these findings as exploratory and leave rigorous measurement of such schema-grounded generalization to future work.

\begin{table*}
  \centering
  \scriptsize
  \begin{tabular}{lccc}
    \hline
    \textbf{Model} & \textbf{Clinton (LLM-EXE)} & \textbf{Bird (LLM-EXE)} & \textbf{CRT-QA (LLM-ACC)} \\
    \hline
    \textbf{o1} & \textbf{59.1\%} & \textbf{25.4\%} & \textbf{69.6\%} \\
    LLaMA Base & 37.4\% & 2.8\% & 41.3\% \\
    LLaMA SFT & \textbf{62.1\%}  \textbf{↑ +24.7} & 1.3\% & 33.7\% \\
    LLaMA SFT-CoT & 56.3\% & 4.7\% & 43.7\% \\
    LLaMA SFT-CoT-GRPO & 57.0\% & \textbf{8\%} \textbf{↑ +5.2} & \textbf{55.3\%} \textbf{↑ +14.0} \\
    Qwen Base & 46.1\% & 11.9\% & 49.0\% \\
    Qwen SFT & \textbf{56.6\%} \textbf{↑ +10.5} & 6.1\% & 45.3\% \\
    Qwen SFT-CoT & 53.6\% & 9.5\% & 45.2\% \\
    Qwen SFT-CoT-GRPO & 52.2\% & \textbf{15.5\%} \textbf{↑ +3.6} & \textbf{56.1\%} \textbf{↑ +7.1} \\
    \hline
  \end{tabular}
\caption{
    \textbf{Performance on Tabular QA and Text-to-SQL.}Performance comparison of OpenAI o1, the 4-bit quantized version of the distilled Deepseek LLaMA
    8B model, and the Qwen-7B-Instruct model (Qwen) evaluated across various datasets. This table compares the
    performance of untrained models (Base), those supervised fine-tuned on the Clinton Dataset (SFT), models fine-
    tuned with Chain-of-Thoughts (SFT-CoT) on the Clinton Dataset, and models that have undergone SFT-CoT on
    the Clinton Dataset and GRPO on the BIRD Dataset. Notable improvements are seen with LLaMA SFT (+24.7\%) and Qwen SFT (+10.5\%) on the Clinton dataset,
    demonstrating the effectiveness of supervised fine-tuning. Our main result is that SFT-CoT-GRPO methodologies enhance
    performance on CRT-QA (tabular reasoning) with gains of +14.0\% and +7.1\%.
  }
\label{tab:model-comparison}
\end{table*}

\subsection{Benefit of CoT Supervision}

Table \ref{tab:model-comparison} reports the performance of our supervised models across Text-to-SQL and tabular question answering tasks. Comparing models fine-tuned with and without CoT supervision, we observe that including reasoning traces slightly reduces performance on the in-domain Clinton dataset, but improves generalization to unseen SQL benchmarks (BIRD) and table-based reasoning tasks (CRT-QA).

We attribute this to the inductive bias introduced by reasoning supervision: models exposed to intermediate inference steps are more likely to learn transferable patterns rather than overfitting to schema-specific templates. Moreover, fine-tuning with CoT traces provides a more structured initialization for reinforcement learning, ensuring that the GRPO stage begins from semantically grounded outputs.

CoT supervision yields markedly different gains for LLaMA and Qwen due to their architectural disparities. In our experiments, a distilled and quantized LLaMA model received a substantially larger performance boost from CoT supervision than the uncompressed Qwen model. We attribute this discrepancy to LLaMA’s compressed nature: distillation and low-precision quantization reduce its representational capacity and can weaken its innate reasoning ability. Consequently, providing explicit step-by-step reasoning guidance during training allows LLaMA to compensate for these lost details, resulting in outsized improvements. In contrast, Qwen—being neither distilled nor quantized—retains a higher precision and fuller pre-trained capacity for reasoning, which means it already performs strongly on complex tasks before CoT fine-tuning. As a result, Qwen's robust baseline reasoning ability leaves less headroom for dramatic gains. This contrast highlights that CoT supervision is especially critical for enhancing compressed models like LLaMA.

\subsection{Text-to-SQL Performance}
The combination of supervised fine-tuning with CoT (SFT-CoT) and GRPO yields marked improvements in Text-to-SQL performance. While the gains on the BIRD dataset—where GRPO was explicitly trained—are anticipated, the enhancement on the Clinton dataset is more notable. This indicates that GRPO not only fine-tunes models to specific tasks but also encourages broader SQL comprehension and reasoning capabilities, facilitating generalization within the Text-to-SQL domain.

In particular, the SFT-CoT + GRPO model shows a strong ability to generalize, demonstrating that models trained on real-world tasks can effectively perform even on data they haven't seen during training, provided they have a strong foundational understanding of SQL reasoning.

\subsection{Zero-shot Question Answering Tabular Reasoning Performance}

Table~\ref{tab:model-comparison} demonstrates that our combined approach of SFT and GRPO, originally fine-tuned on Text-to-SQL data, also enhances tabular reasoning performance in zero-shot settings. Specifically, when evaluated on CRT-QA, we observe improved reasoning across the model, showcasing that the model’s exposure to SQL structures helps it tackle general tabular question answering tasks even when SQL generation is not explicitly required.

The zero-shot performance is indicative of the transferability of the reasoning skills learned during SQL task training. By implicitly learning to reason over structured tables in the SQL framework, the model becomes better at navigating more complex question answering tasks, further underlining the value of using SQL as a foundational tool for structured data reasoning.

\subsection{Reward Ablation}

In this section, we investigate the contribution of various reward functions in our GRPO training. Table~\ref{tab:ablation reward} presents the results of our ablation study, evaluating the impact of different reward configurations on the model’s performance on the BIRD and CRT-QA tasks. Specifically, we analyze the effect of different combinations of rewards—including execution-based, string matching, component-level matching, and LLM-based judgment rewards—on the accuracy of SQL execution and tabular question answering.

Ablation studies indicate that string matching serves as the most effective single reward due to its continuous nature, facilitating initial learning. However, exclusive reliance on string matching can lead to diminished performance in later training stages. We observe that combining string matching with additional reward mechanisms enhances overall effectiveness, as the initial continuous reward provides a substantial learning advantage. The most promising two-reward combination identified is string matching coupled with the LLM Judge Reward with classes. This synergistic approach effectively merges the continuous evaluation of string accuracy with the discrete assessment of general SQL quality, thereby creating a robust framework for improved model performance.

From the results in Table~\ref{tab:ablation reward}, we observe that incorporating a broader range of reward functions generally improves model performance. For instance, the best four rewards configuration shows significant improvements on CRT-QA for the LLaMA model, indicating that a more diverse set of feedback signals enhances generalization across tasks. This suggests that combining different reward signals allows the model to better capture both syntactic correctness (in SQL) and logical coherence (in tabular reasoning), leading to a more balanced and accurate reasoning process.

\begin{table}
  \centering
  \scriptsize
  \begin{tabular}{l@{\hspace{2mm}}c@{\hspace{2mm}}c}
    \hline
    \textbf{Reward Configuration} & \textbf{BIRD (LLM-EXE)} & \textbf{CRT-QA (LLM-ACC)} \\
    \hline
    \textbf{Best Reward (LLaMA)} & 8.2\% & 51.7\% \\
    \textbf{Best Reward (Qwen)} & 9.1\% & 54.3\% \\
    \textbf{Best Two Rewards (LLaMA)} & 8.6\% & 54.8\% \\
    \textbf{Best Two Rewards (Qwen)} & 10.8\% & 55.0\% \\
    \textbf{Best Four Rewards (LLaMA)} & \textbf{8.0\%} & \textbf{55.3\%} \\
    \textbf{Best Four Rewards (Qwen)} & \textbf{15.5\%} & \textbf{56.1\%} \\
    \hline
  \end{tabular}
  \caption{
    \textbf{Ablation study of reward configurations.} The models initially underwent SFT on Chain-of-Thought traces on Clinton, followed by GRPO on BIRD, where specific reward functions were applied.  Performance is evaluated across the best reward configurations: the best reward is derived from string matching, the best two consist of both string matching and evaluations by an LLM judge, and the best four are based on a combination of string matching, component-level matching, execution-based reward, and evaluations by an LLM judge. Evaluation scores include execution accuracy (LLM-EXE) and accuracy (LLM-ACC) assessed by an LLM judge. 
  }
  \label{tab:ablation reward}
\end{table}

\section{Conclusion}

Our experiments demonstrate that treating Text-to-SQL as a structured reasoning task—rather than just query generation—significantly enhances LLMs’ ability to understand tabular data. We use a two-stage framework:

\begin{enumerate}
    \item \textbf{Synthetic Chain-of-Thought Supervision}: generates detailed, clause-level reasoning traces to teach models how to traverse, filter, and aggregate table fields.
    
    \item \textbf{Reinforcement Learning with GRPO}: leverages multiple reward signals to align SQL execution accuracy with broader, transferable reasoning behaviors.
\end{enumerate}

Our work goes beyond execution optimization by integrating multi-dimensional reward signals and leveraging the effects of GRPO for broader tabular reasoning tasks, such as question answering (QA). In this way, GRPO is not just an optimizer but a vehicle for generalization and reasoning alignment. The distilled and quantized LLaMA model achieves a relative 33.9\% increase, while the Qwen model sees a relative 14.5\% improvement on the tabular reasoning CRT-QA benchmark.

These results confirm that SQL serves not only as a task-specific formalism but also as an effective scaffold for teaching robust, transferable reasoning over structured data. Moving forward, we plan to scale this method to larger models, explore richer reward configurations, and extend the framework to other formal languages and real-world datasets, aiming to build more interpretable and capable systems for structured data interaction.

\section*{Limitations}

Our study focuses on medium-scale foundation models—distilled LLaMA and Qwen-7B—whose exact pretraining corpora are undocumented. As a result, we cannot determine coverage or gaps across domains, languages, or proprietary material. This opacity complicates any analysis of domain blind spots, spurious correlations, or memorization risks. Moreover, the relatively modest parameter counts of these models may limit performance on tasks requiring deep domain expertise, such as biomedical or legal reasoning.

We evaluate tabular reasoning using CRT-QA, with o3-mini serving as an automated judge. While expedient, this setup lacks the nuance of human evaluation, particularly for complex reasoning and semantic alignment. Additionally, standard Text-to-SQL and tabular QA benchmarks may under-represent the ambiguity and noise present in real-world data, making our results more indicative of structured reasoning progress than deployment readiness.

Our current framework employs only two training stages. In contrast, multi-phase pipelines such as R1 leverage up to four stages, including instruction tuning and iterative CoT refinement. While our approach prioritizes simplicity and efficiency, it may sacrifice opportunities for deeper alignment or curriculum structuring.

Future research should address these limitations by exploring larger, better-documented models, human-in-the-loop evaluation, and more diverse datasets. Additional training stages—such as pre-CoT bootstrapping or domain-adaptive pretraining—may further enhance generalization and robustness in real-world table reasoning.

\bibliographystyle{plain} 
\bibliography{bib} 

\appendix

\section{Summary of Clinton Dataset} \label{app:clinton}

We conduct part of our evaluation using the \texttt{Clinton/Text-to-sql-v1} dataset,\footnote{https://huggingface.co/datasets/Clinton/Text-to-sql-v1} a large-scale compilation of natural language to SQL examples spanning a broad set of domains. This benchmark includes 26 individual datasets, covering academic records, medical databases, entertainment metadata, government statistics, and more.

Each example in the dataset consists of a natural language query, an associated database schema, and a corresponding SQL statement. Some subsets also include table content or ground-truth execution results. The diversity in schema complexity and domain coverage makes this benchmark well-suited for evaluating both generalization and transfer in Text-to-SQL and tabular reasoning models.

Key datasets include:
\begin{itemize}
    \item \textbf{Spider} \cite{yu2018spider} – Complex, cross-domain Text-to-SQL benchmark.
    \item \textbf{WikiSQL} \cite{zhong2017seq2sql} – Large-scale dataset with simple queries over Wikipedia tables.
    \item \textbf{ATIS} \cite{hemphill1990atis} – Airline travel information with traditional semantic parsing annotations.
    \item \textbf{MIMICSQL} \cite{wang2020text} and \textbf{eICU} \cite{pollard2018eicu} – Clinical databases for medical question answering.
\end{itemize}

We also include lesser-known and synthetic datasets such as \texttt{Criteria2SQL} \cite{fang2022combining}, \texttt{SEDE} \cite{hazoom2021text}, \texttt{SQuALL} \cite{shi2020potential}, and \texttt{NVBench} \cite{wang2023natural}, along with public domain tabular corpora like IMDb, Yelp, and historical sports or wildfire datasets.

This variety allows us to test the ability of LLMs to reason across database schemas, interact with realistic tabular structures, and generalize beyond fixed SQL templates.

\section{Prompts} \label{app:prompt}

\subsection{Creating Synthetic CoT} \label{app:create_Sythetic_COT}

This section outlines the structure of prompts designed for SQL query generation tasks. Each prompt features SQL table schemas and clear instructions, facilitating the generation of valid SQL queries using SQLite syntax. The expert guidance within the prompts emphasizes the requirement to articulate the reasoning behind the constructed SQL queries. By utilizing this approach, we aim to train models that can effectively understand the context of relational data and generate precise queries that meet specific operational goals, thereby enhancing the overall interpretability and accuracy of automated SQL generation.

\begin{center}
    \fbox{\parbox{\columnwidth}{\scriptsize
     You are a SQL expert. Below are SQL table schemas paired with instructions that describe a specific task. Using valid SQLite syntax, write a response that appropriately completes the request for the provided tables.

    \textbf{SCHEMA: {schema}}
    
    \textbf{INSTRUCTIONS: {specific task instructions}}
    
    When answering, provide reasoning for the SQL query you create using the following template:
    
    <sql> Write the SQL query here, ensuring it adheres to SQLite syntax and effectively accomplishes the task described in the instructions. </sql>
    }}
\end{center}

\subsection{Evaluation of Synthetic CoT} \label{app:eval_Sythetic_COT}

This section specifies a prompt for evaluating the correctness of SQL queries based on a defined schema and a reference SQL query. The prompt clearly delineates the evaluation task for the SQL expert, presenting the query to be evaluated, the relevant schema, and the correct SQL reference. The evaluator is instructed to determine whether the provided SQL query is correct or incorrect, with responses limited to "Correct" or "Wrong." This structured approach facilitates precise assessment of SQL queries, contributing to the development of robust models capable of generating and validating SQL syntax effectively.

\begin{center}
    \fbox{\parbox{\columnwidth}{\scriptsize
        You are an SQL expert, and your task is to evaluate whether the SQL query below is correct based on the provided schema and the correct SQL reference.
        
        \textbf{SQL Query:} {ans.sql}
        
        \textbf{Schema:} {schema}
        
        \textbf{Correct SQL:} {correct\_sql}
        
        Return ONLY "Correct" or "Wrong".
    }}
\end{center}

\subsection{LLM Judge for Execution Based Reward}
\label{app:llm_exe_prompt}

For our Execution Reward in Group Relative Policy Optimization (GRPO) the LLM judge is instructed to count the number of orthographic changes required to convert each predicted query into the corresponding correct query. The reward is computed using the following equation:

\begin{equation} R_{\text{exec}} = \frac{1}{x + 1}, \end{equation} where $x$ is the number of detected changes. This methodology provides a more continuous measure of execution accuracy, crucial for refining the model's performance.

\begin{center}
    \fbox{\parbox{\columnwidth}{\scriptsize
You are an SQL expert. Count how many changes you need to make to get the following predicted queries correct.
    
\textbf{Predicted Queries (one per line)}: 
{queries\_to\_rank}

\textbf{For reference, use this Schema:} {schema}.
    
\textbf{Here is the correct query}: 
{true\_query}
    
You should count the number of Orthographic elements you need to change from the predicted queries to the correct query.
    
ONLY RETURN a JSON object with a single 'scores' field containing a list of \textbf{num\_queries} numbers reflecting the number of changes needed for each predicted query.
    }}
\end{center}

\subsection{LLM Judge with Classes}
\label{app:llm_classes_prompt}

The LLM judge reward is designed to evaluate the quality of predicted SQL queries by comparing them to a reference correct query. In this task, the judge is instructed to assign a grade to each predicted query on a scale from 'Very bad' to 'Excellent.' The grading criteria are explicitly defined, allowing the judge to assess various aspects of the queries, including grammatical correctness, logical accuracy, and overall fidelity to the correct query. This structured grading system enables a nuanced analysis of the model's output quality, providing insights that facilitate targeted improvements in query generation. 

\begin{center}
    \fbox{\parbox{\columnwidth}{\scriptsize
    Compare these SQL queries to the correct query and grade each one as: 'Very bad', 'Bad', 'Above average', 'Good', or 'Excellent'.
    Use the following grading system, and the correct query as reference :
    
    \textbf{Correct Query:} {true\_query}
    
    \textbf{1. Excellent:} this is only given when the SQL query is perfect and matches \{true\_query\}
    
    \textbf{2. Good:} This is when there is a grammar mistake in the query
    
    \textbf{3. Above average:} This is when the query is mostly correct but gets a logical step wrong in the query
    
    \textbf{4. Bad:} Makes more than one mistake in the query 
    
    \textbf{5. Very bad:} does not produce a query or varies significantly from the correct query

    \textbf{Queries to grade:}
    {queries\_to\_rank}

    \{format\_instructions\}
    }}
\end{center}

\section{Implementation Details} \label{app:implementation}

We used the \texttt{Unsloth} framework\footnote{\url{https://github.com/unslothai/unsloth}} for efficient fine-tuning of LLMs. Unsloth provides support for QLoRA-style training with Flash Attention 2, bitsandbytes quantization, and PEFT-compatible adapters.

We fine-tuned two pretrained models:
\begin{itemize}
    \item \textbf{Qwen-16B}, a dense, instruction-tuned model released by Alibaba DAMO, trained in full precision.
    \item \textbf{DeepSeek LLaMA3-8B}, a 4-bit quantized variant of Meta’s LLaMA 3–8B, optimized and released by DeepSeek AI\footnote{\url{https://huggingface.co/deepseek-ai/deepseek-llm-8b-base}}.
\end{itemize}

Supervised fine-tuning (SFT) was performed on the Clintondataset using QLoRA adapters, while reinforcement learning with GRPO was applied on the BIRD benchmark. The GRPO setup used candidate comparisons and execution-guided rewards computed via SQLite.

Experiments were conducted on 4×A100 80GB GPUs using mixed-precision (FP16).

\end{document}